\documentclass[]{ailab}

\usepackage{tcolorbox}
\usepackage{fontawesome}
\usepackage{hyperref}
\usepackage{wrapfig}
\usepackage{tabularx}
\usepackage{textcomp}
\usepackage{stfloats}
\usepackage{url}
\usepackage{verbatim}
\usepackage{titlesec}
\usepackage{adjustbox}
\usepackage{multirow}
\usepackage{pifont}
\usepackage{tikz}
\usepackage{comment}
\usepackage{amsmath,amssymb}
\usepackage{colortbl}
\usepackage{color}
\usepackage{booktabs}
\usepackage{natbib}
\setcitestyle{square,comma,numbers,sort&compress}
\usepackage{graphicx}
\usepackage{subcaption}
\usepackage[dvipsnames]{xcolor}
\usepackage{algorithm}
\usepackage{listings}
\usepackage{lineno}
\usepackage{enumitem}
\usepackage{float}
\usepackage[normalem]{ulem}
\usepackage{diagbox}
\usepackage{makecell}
\usepackage{array}
\usepackage{rotating}

\RequirePackage{xspace}
\makeatletter
\DeclareRobustCommand\onedot{\futurelet\@let@token\@onedot}
\def\@onedot{\ifx\@let@token.\else.\null\fi\xspace}

\definecolor{darkblue}{rgb}{0, 0, 0.5}
\hypersetup{colorlinks=true, citecolor=darkblue, linkcolor=darkblue, urlcolor=darkblue}

\definecolor{codeblue}{rgb}{0.25, 0.5, 0.5}
\definecolor{codekw}{rgb}{0.35, 0.35, 0.75}
\lstdefinestyle{Pytorch}{
    language = Python,
    backgroundcolor = \color{white},
    basicstyle = \fontsize{9pt}{8pt}\selectfont\ttfamily\bfseries,
    columns = fullflexible,
    aboveskip=1pt,
    belowskip=1pt,
    breaklines = true,
    captionpos = b,
    commentstyle = \color{codeblue},
    keywordstyle = \color{codekw},
}

\usepackage{amsmath,amsfonts,bm}

\def\1{\bm{1}}

\makeatother

\newlength\savewidth

\newcolumntype{x}[1]{>{\centering\arraybackslash}p{#1pt}}
\newcolumntype{y}[1]{>{\raggedright\arraybackslash}p{#1pt}}
\newcolumntype{z}[1]{>{\raggedleft\arraybackslash}p{#1pt}}

\makeatletter
\newcommand{\thickhline}{%
 \noalign {\ifnum 0=`}\fi \hrule height 1pt
 \futurelet \reserved@a \@xhline
}
\makeatother

\definecolor{green}{HTML}{009000}
\definecolor{red}{HTML}{ea4335}

\renewcommand{\paragraph}[1]{\vspace{1.25mm}\noindent\textbf{#1}}

\title{\centering \textbf{FlashMemory-DeepSeek-V4: Lightning Index Ultra-Long Context via Lookahead Sparse Attention}}

\author[1, *, \dagger]{Yan Wang}
\author[2, 3, *]{Qifan Zhang}
\author[2, 4, *]{Jiachen Yu}
\author[2, *]{Tian Liang}
\author[1, *]{Dongyang Ma}
\author[2]{Xiang Hu}
\author[2]{Zibo Lin}
\author[2]{Chunyang Li}
\author[2]{Zhichao Wang}
\author[2, 3]{Miao Peng}
\author[2]{Nuo Chen}
\author[3]{Jia Li}
\author[4]{Yujiu Yang}
\author[2]{Haitao Mi}
\author[2]{Dong Yu}

\affiliation[1]{Independent Researchers\\}
\affiliation[2]{Tencent}
\affiliation[3]{The Hong Kong University of Science and Technology (Guangzhou)\\}
\affiliation[4]{Tsinghua University}
\contribution[*]{Equal contribution}
\contribution[\dagger]{Project Lead}

\email{yanwang.branden@gmail.com}

\abstract{
Conventional LLMs keep the full KV cache loaded during decoding, causing a severe GPU memory bottleneck for ultra-long context serving. In this report, we propose \textbf{Lookahead Sparse Attention (LSA)}, a novel inference paradigm powered by a Neural Memory Indexer built upon the DeepSeek-V4 architecture. Rather than passively attending to all historical tokens, LSA proactively predicts future context demands and preserves only the query-critical KV chunks in the GPU memory. Crucially, we instantiate this architecture via a \textbf{backbone-free decoupled training} strategy. By formulating the indexer as a standard dual-encoder architecture, we train it independently using standard retrieval training frameworks without ever loading the massive backbone model into GPU memory. 

We demonstrate that this ``less is more'' paradigm significantly maximizes serving efficiency while acting as an effective attention denoiser in tasks that rely on long-term global memory. Across primary long-context evaluation suites (e.g., LongBench-v2, LongMemEval, and RULER), \texttt{FM-DS-V4} compresses the average physical KV cache footprint down to merely 13.5\% of the full-context baseline, while consistently preserving or slightly elevating downstream accuracy (+0.6\% absolute margin on average). At 1M context, per-decode-token compute drops to 0.30$\times$ of the baseline and GPU KV cache shrinks by 90\% (3.73$\to$0.37 GB), translating into \textbf{2.8$\times$ aggregate throughput and 2.7$\times$ concurrency gains} in PD-disaggregated serving on 8$\times$H20 GPUs.
}

\headercontent{
    \begin{tabular}{cc}
        \href{https://github.com/libertywing/FlashMemory-Deepseek-V4}{\faGithub\ Code} &
        \hspace{2em}
        \href{https://huggingface.co/libertywing/FlashMemory-Deepseek-V4}{%
            \raisebox{-0.15\height}{\includegraphics[height=1.1em]{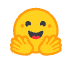}}\ Model%
        }
    \end{tabular}
}

\begin{document}
\thispagestyle{firstheader}
\maketitle

% ==================== PROJECT STATUS NOTICE BOX ====================
\vspace{-0.2cm}
\begin{center}
\definecolor{noticeblue}{RGB}{235, 243, 250}
\definecolor{borderblue}{RGB}{70, 130, 180}
\begin{tcolorbox}[
    colback=noticeblue,
    colframe=borderblue,
    arc=4px,
    boxrule=0.8pt,
    left=6pt, right=6pt, top=3pt, bottom=3pt,
    width=\linewidth
]
    \small
    \textbf{Project Status:} \\
    Due to organizational realignments, the Project Lead has parted ways with Tencent, and this project has been suspended. 
    We firmly believe in the potential of the \textit{LSA} paradigm for infinite long-context intelligence. If you or your organization are interested in supporting or collaborating on the next phase (e.g., compute sponsorship, scaling tests, or research integration), please contact the Project Lead at \href{mailto:yanwang.branden@gmail.com}{\texttt{yanwang.branden@gmail.com}}.
\end{tcolorbox}
\end{center}
\vspace{0.2cm}
% =====================================================================
% ==================== FIRST PAGE TEASER FIGURE ====================
\begin{figure}[H]
    \centering
    \includegraphics[width=0.92\linewidth]{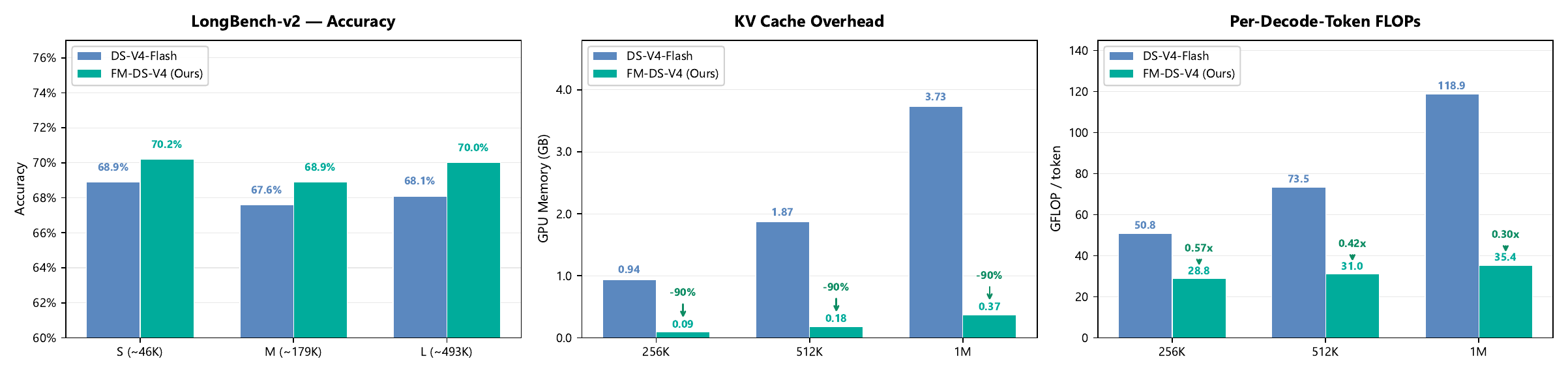}
    \vspace{-6pt}
    \caption{\textbf{KV cache and per-token FLOPs reductions deliver 2.8$\times$ throughput and 2.7$\times$ concurrency gains at 1M context.} On LongBench-v2, FM-DS-V4 consistently matches or exceeds \texttt{DS-V4-Flash}. At 1M context, per-decode-token compute drops from 118.9 to 35.4 GFLOP (0.30$\times$) while GPU KV cache shrinks by 90\% (3.73$\to$0.37 GB). KV cache Overhead and FLOPs are measured via sglang deployment logs on an 8$\times$H20 GPU server.}
    \label{fig:teaser_joint}
\end{figure}
\vspace{-0.3cm}
% =====================================================================

\section{Introduction}
\label{sec:Introduction}

The extension of Large Language Models (LLMs) toward ultra-long context windows is fundamentally bottlenecked by memory capacity. While modern sparse attention mechanisms successfully reduce the computational FLOPs per decoding step to a near-constant level, the GPU memory footprint of the Key-Value (KV) cache still scales linearly with the sequence length. Recent foundation models like DeepSeek-V4\footnote{\url{https://huggingface.co/deepseek-ai/DeepSeek-V4-Pro}} and Qwen3.5\footnote{\url{https://huggingface.co/Qwen/Qwen3.5-397B-A17B}} attempt to slow down this memory explosion by incorporating heavily compressed attention (HCA) or linear attention layers \cite{deepseekv4, qwen35blog}. However, to preserve fine-grained factual recall, these models must still retain a significant portion of low-compression or full-attention layers \cite{deepseekv4}. Consequently, they only mitigate the rate of memory growth rather than eliminating the linear scaling bottleneck itself.

This work stems from a simple yet striking observation of resource waste during inference: conventional LLMs fully load and carry the entire KV cache in GPU memory even when the active decoding step is completely independent of the historical context. Our empirical analysis of real-world inference logs reveals that \textbf{over 90\% of user requests with contexts longer than 64K tokens can be accurately resolved using only the last 8K tokens.} This indicates that an overwhelming majority of GPU memory is squandered on inactive context that contributes nothing to the current token prediction. Conversely, simply discarding history via standard sliding-window attention fails entirely on the remaining tasks that genuinely require global context synthesis. This hard contradiction---supporting deep global reasoning without paying the full GPU memory tax for local generation steps---is the root cause behind the prohibitive cost of long-context serving.

To resolve this dilemma, we present \textbf{Lookahead Sparse Attention (LSA)}. Following the structural compression spirit of DeepSeek-V4 \cite{deepseekv4}, our architecture retains all highly condensed HCA chunks (128:1 compression ratio) to maintain global context awareness. However, we fundamentally upgrade the conventional Compressed Sparse Attention (CSA) layers into our predictive LSA paradigm. LSA empowers the model to \textit{not recall that much} fine-grained context; instead, driven by a highly efficient \textbf{Neural Memory Indexer}, the system triggers periodically at a fixed decoding interval of $\tau$ steps (e.g., $\tau = 64$) to evaluate current hidden states and proactively fetch only the critical CSA chunks into the GPU memory. Crucially, we formulate the indexer as a standalone dual-encoder architecture. This decoupled design allows us to train the indexer independently on pre-computed hidden states and labels, completely bypassing the prohibitive memory and computational overhead of full-model fine-tuning or joint distillation. 

Experimental results across three distinct long-context benchmarks confirm the robustness and striking efficiency of LSA. In scenarios requiring long-term memory and deep understanding, LSA acts as an effective attention denoiser. Specifically, averaged across LongBench-v2, LongMemEval, and RULER, \texttt{FM-DS-V4} compresses the average physical KV cache footprint down to merely 13.5\% of the full-context baseline, while consistently preserving or slightly elevating downstream accuracy (+0.6\% absolute margin on average). At 1M context, per-decode-token compute drops to 0.30$\times$ of the baseline and GPU KV cache shrinks by 90\% (3.73$\to$0.37 GB), translating into \textbf{2.8$\times$ aggregate throughput and 2.7$\times$ concurrency gains} in PD-disaggregated serving on 8$\times$H20 GPUs.

In summary, our core contributions are threefold:
\begin{itemize}
    \item \textbf{Lookahead Sparse Attention (LSA) Paradigm:} We propose LSA, a novel inference paradigm that eliminates the hard contradiction between long-context modeling capabilities and hardware efficiency by proactively predicting and fetching query-critical KV chunks on demand.
    \item \textbf{Backbone-Free Decoupled Training:} We introduce an ultra-lightweight training strategy that physically isolates the indexer from the host LLM. Formulated as a standalone dual-encoder trained on pre-computed representations, the indexer can be optimized independently in just \textbf{a single H20 GPU hour} without ever loading the massive backbone model.
    \item \textbf{Breakthrough in Serving Efficiency:} LSA reduces average GPU KV cache to merely 13.5\% of the baseline while maintaining accuracy. At 1M context, this translates to a 0.30$\times$ per-token compute ratio and 90\% KV cache reduction, yielding \textbf{2.8$\times$ aggregate throughput and 2.7$\times$ concurrency gains} in PD-disaggregated serving on 8$\times$H20 GPUs.
\end{itemize}

\begin{figure}[t]
\centering
\includegraphics[width=\linewidth]{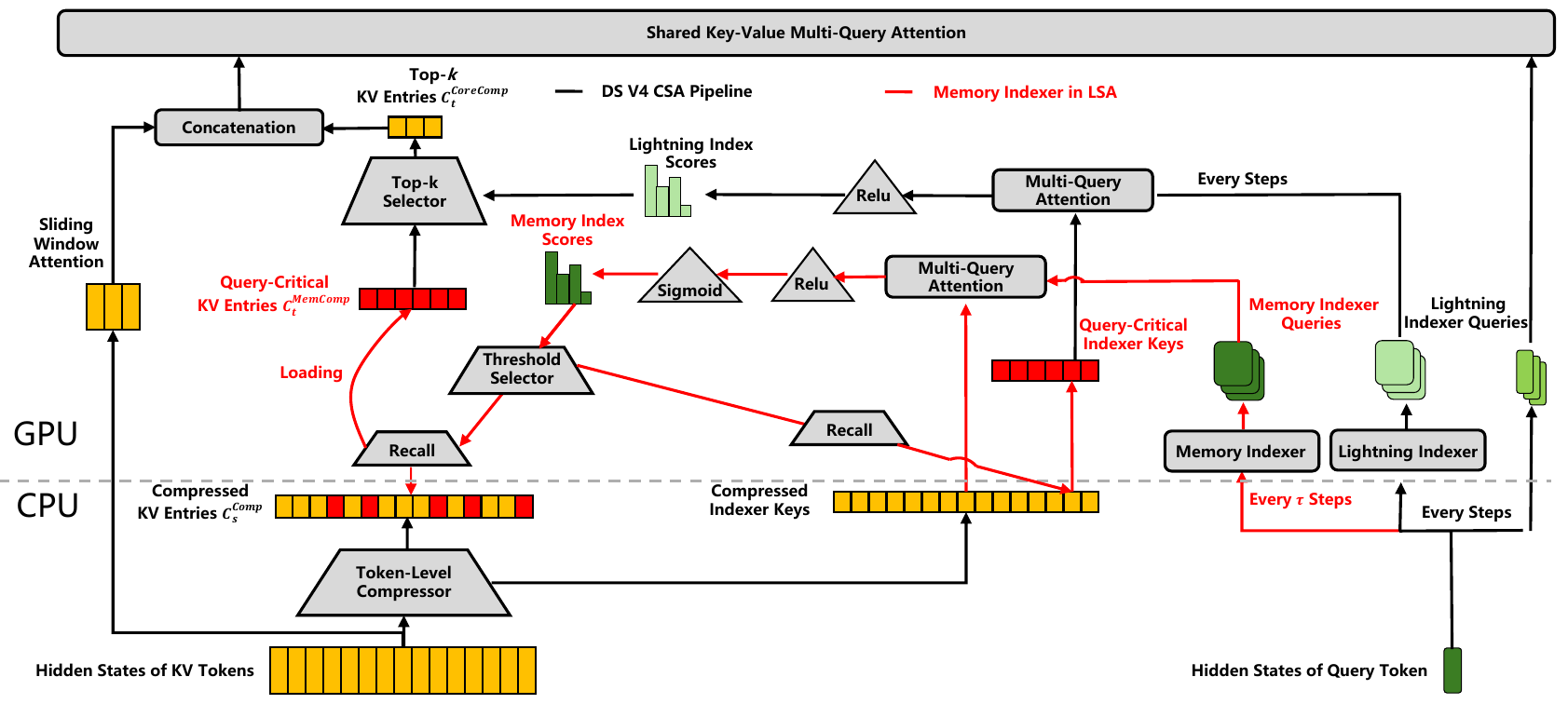}
\caption{Architectural overview of LSA vs. CSA. The black lines denote the standard, step-by-step CSA pipelines. The red lines highlight our proposed LSA mechanism, which decouples the GPU memory footprint by leveraging a Memory Indexer to fetch historical KV chunks dynamically every $\tau$ steps.}
\label{fig:arch}\label{fig:intro}
\end{figure}

\section{Methodology}
\label{sec:methodology}

In this section, we present the technical details of Lookahead Sparse Attention (LSA), including its architectural formulation, data curation pipeline, optimization strategy, optimal configuration, and serving infrastructure. 
Specifically, Section~\ref{subsec:architecture} introduces how we architect LSA on top of the DeepSeek-V4 framework to achieve predictive context selection. 
Section~\ref{subsec:data_construction} introduces our lookahead data formats and the automated gathering pipeline. 
Section~\ref{subsec:training} details our decoupled training strategy that physically isolates indexer optimization from the massive LLM backbone.
Section~\ref{subsec:optimal_config} presents our systematic exploration of the optimal layer configuration and training recipe for the production model.
Finally, Section~\ref{sec:infra} describes how LSA is deployed in a PD-disaggregated serving stack, including the CPU--GPU memory split that realizes the algorithmic sparsity as hardware savings.

\subsection{Memory Indexer for Lookahead Selection}
\label{subsec:architecture}

The core design principle of LSA is to minimize modifications to the DeepSeek-V4 architecture, thereby maximizing the preservation of its established capabilities. Therefore, our Memory Indexer mirrors the exact architecture of the native Lightning Indexer used in DeepSeek-V4, reusing the compressed indexer keys $K^{\text{IComp}}$ as the dense representation of historical context. The definitive departure is that we introduce a Sigmoid function as the final activation layer to scale the indexer scores into the $(0,1)$ range, and we replace the rigid Top-$k$ selector with a threshold-based mechanism to recall a dynamic number of historical entries.

During the autoregressive decoding stage, the Memory Indexer triggers periodically at a fixed decoding step interval $\tau$ (e.g., $\tau = 64$) to perform lookahead block prediction. As illustrated in Figure~\ref{fig:intro}, at decoding step $t$ (where $t \pmod \tau = 0$), given the current input hidden state of the query token $\mathbf{h}_t \in \mathbb{R}^d$, we map it into low-rank indexer queries across $n_h^l$ indexer heads:
\begin{align}
    \mathbf{c}_t^Q &= \mathbf{h}_t \cdot W^{DQ}, \\
    [\mathbf{q}_{t,1}^l; \mathbf{q}_{t,2}^l; \dots; \mathbf{q}_{t,n_h^l}^l] = \mathbf{q}_t^l &= \mathbf{c}_t^Q \cdot W^{IUQ},
\end{align}
where $W^{DQ} \in \mathbb{R}^{d \times d_c}$ and $W^{IUQ} \in \mathbb{R}^{d_c \times c^l n_h^l}$ represent the down-projection and up-projection matrices for the lookahead query representation, respectively. Concurrently, we dynamically project $\mathbf{h}_t$ to compute the routing head weights $\mathbf{w}_t^l$:
\begin{equation}
    [\mathbf{w}_{t,1}^l; \mathbf{w}_{t,2}^l; \dots; \mathbf{w}_{t,n_h^l}^l] = \mathbf{w}_t^l = \mathbf{h}_t \cdot W^w,
\end{equation}
where $W^w \in \mathbb{R}^{d \times n_h^l}$ is a learnable matrix, and $\mathbf{w}_{t,h}^l$ dynamically scales the importance of the $h$-th indexer head. 

To determine which historical compressed KV entries are strictly critical for the upcoming window $[t, t+\tau-1]$, the lookahead index score $I_{t,s}$ between the query token $t$ and a preceding compressed entry $s$ ($s < \lfloor\frac{t}{m}\rfloor$) is formulated as a head-fused gated matching score with a Sigmoid activation:
\begin{equation}
    I_{t,s} = \sigma \left( \sum_{h=1}^{n_h^l} \mathbf{w}_{t,h}^l \cdot \text{ReLU}\left( \mathbf{q}_{t,h}^l \cdot \left(K_s^{\text{IComp}}\right)^T \right) \right),
\end{equation}
where $\sigma(\cdot)$ denotes the standard Sigmoid function. 

This Sigmoid activation stands as the only architectural departure from the native Lightning Indexer. While the original one applies a ReLU boundary for raw attention scoring, LSA introduces Sigmoid normalization to align the Memory Indexer's scalar outputs explicitly with discrete binary targets $y \in \{0, 1\}$. For a query token $t$, rather than a rigid Top-$k$ selection strategy, we fetch all preceding compressed KV entries whose lookahead scores meet or exceed a specific classification threshold (i.e., $I_{t,s} \ge 0.5$) from the CPU Cold Pool into the GPU memory for subsequent core attention (the physical realization of this two-tier CPU--GPU transfer is detailed in Section~\ref{sec:infra}):
\begin{equation}
    \label{eq:memcomp}
    C_t^{\text{MemComp}} = \left\{ C_s^{\text{Comp}} \;\middle|\; I_{t,s} \ge 0.5 \right\},
\end{equation}
where $C^{\text{Comp}}$ denotes the pre-computed compressed KV entries. Once the query-critical context subset $C_t^{\text{MemComp}}$ is successfully resident in the GPU memory, the native Lightning Indexer calculates the token-level matching scores within this restricted $C_t^{\text{MemComp}}$ boundary instead of scanning the full context. It applies the native ReLU-based Multi-Query Attention scoring over the fetched subset to select the final fine-grained Top-$k$ core compressed entries:

\begin{equation}
    \label{eq:corecomp}
    C_i^{\text{CoreComp}} = \left\{ C_s^{\text{Comp}} \in C_t^{\text{MemComp}} \;\middle|\; \text{Score}_{\text{native}}(i, s) \in \text{Top-}k \right\}.
\end{equation}
The selected $C_i^{\text{CoreComp}}$ entries are then concatenated with the non-offloadable sliding window KV cache to participate in the final core attention computation. This tiered selection mechanism guarantees that the underlying FlashInfer or FlashAttention kernels operate exclusively on a highly condensed, hardware-resident active sequence footprint, enabling the physical eviction of all non-recalled KV entries to the CPU (Section~\ref{sec:infra}).

\subsection{Lookahead Dataset Construction}
\label{subsec:data_construction}

The cornerstone of optimizing our Memory Indexer is pinning down exactly which historical compressed KV entries a decoding token needs to look ahead to. 
A naive approach would define the positive label set for token $t$ as the simple union of all Top-$k$ entries recalled by the native Lightning Indexer across the future window $[t, t+\tau-1]$. 
However, empirical analysis reveals a massive inflation problem with this strategy, resulting in nearly 10,000 positive samples per token window before filtering (reduced to approximately 100--1,000 after our pipeline).
The root cause is that a rigid Top-$k$ selector forces the model to recall a fixed number of preceding entries regardless of their actual relevance, causing low-probability noise entries from different attention layers to heavily pollute the ground-truth dataset.

To eliminate this noise, we propose an golden label filtering pipeline that uses a \textbf{Cross-Layer Majority Voting} mechanism to identify the true ``golden entries.'' 
The data generation pass runs completely offline on the frozen DeepSeek-V4-Flash backbone model. 
For each decoding token $i \in [t, t+\tau-1]$ and across all $L$ CSA layers (where $L=21$ for DeepSeek-V4-Flash \cite{deepseekv4}), we extract the raw indexer logit scores $S_{i,l,s}$ for every preceding compressed entry $s$. 
We then filter these scores through a three-step denoising pipeline:

\begin{itemize}
    \item \textbf{Step 1: Softmax Normalization.} We convert the raw logit scores into a valid probability distribution via a Softmax operation over all historical entries:
    \begin{equation}
        P_{i,l,s} = \frac{\exp(S_{i,l,s})}{\sum_{j} \exp(S_{i,l,j})}.
    \end{equation}
    
    \item \textbf{Step 2: Top-$p$ Thresholding.} Instead of using a fixed Top-$k$ count, we dynamically retain only the high-confidence entries using a nucleus threshold $p$ (we empirically set $p=0.6$). An entry $s$ is marked as selected by layer $l$ if it falls within the minimum set of entries that cumulatively account for the top $60\%$ of the probability mass:
    \begin{equation}
        \mathcal{M}_{i,l} = \left\{ s \;\middle|\; \sum_{j \in \text{Sorted}(P_{i,l,:})} P_{i,l,j} \le p \right\}.
    \end{equation}
    
    \item \textbf{Step 3: Cross-Layer Majority Voting.} We aggregate the selection hits across all $L$ layers. The voting score $V_{i,s}$ for entry $s$ at token step $i$ is calculated by counting how many layers independently voted for it:
    \begin{equation}
        V_{i,s} = \sum_{l=1}^{L} \mathbb{I}(s \in \mathcal{M}_{i,l}),
    \end{equation}
    where $\mathbb{I}(\cdot)$ is the indicator function. An entry is officially recognized as a core active entry $\mathcal{A}_i^{\text{golden}}$ if and only if it secures consensus backing from at least $\theta$ layers (we set $\theta = 3$):
    \begin{equation}
        \mathcal{A}_i^{\text{golden}} = \left\{ s \;\middle|\; V_{i,s} \ge 3 \right\}.
    \end{equation}
\end{itemize}

Finally, for each lookahead evaluation window triggered at decoding step $t$, the positive ground-truth label set $\mathcal{Y}_t^{+}$ is established by taking the union of these denoised golden entries across the entire future temporal window of $\tau$ steps:
\begin{equation}
    \label{eq:label}
    \mathcal{Y}_t^{+} = \bigcup_{i=t}^{t+\tau-1} \mathcal{A}_i^{\text{golden}}.
\end{equation}

By shifting from an arbitrary Top-$k$ lookup to a consensus-driven density estimation, our pipeline isolates the true contextual backbone of the long sequence, discarding irrelevant background noise. In total, our training set comprises approximately 10,000 long documents with context lengths ranging from 16K to 512K tokens.

\subsection{Optimization and Decoupled Training}
\label{subsec:training}

Although our Memory Indexer shares a structural setup similar to the native Lightning Indexer, their underlying optimization paradigms are fundamentally different. 
Unlike the native Lightning Indexer which relies on heavy end-to-end self-distillation, we treat the Memory Indexer as a standard retrieval model and optimize it via \textbf{metric learning}. 
The primary training objective is to perform distance-based contrastive optimization: maximizing the lookahead matching scores for query-critical historical entries while minimizing the scores for negative samples.

A key system insight of LSA is that the compressed indexer keys $K^{\text{IComp}}_s$ of historical entries are entirely pre-computed and strictly \textbf{frozen} during the training stage. 
Consequently, the optimization process simplifies into training only the query encoder of a standard dual-encoder retrieval architecture. 
Specifically, we only need to optimize the low-rank projection matrices ($W^{DQ}, W^{IUQ}, W^w$) to map the current input hidden state $\mathbf{h}_t$ to align with the fixed historical targets. 

To achieve this objective, we minimize a standard element-wise Binary Cross-Entropy (BCE) loss function over the predicted lookahead scores. For a single sample with predicted probability $p$ and label $y \in \{0,1\}$, the per-sample BCE is defined as:
\begin{equation}
    \ell_{\text{BCE}}(p, y) = - \bigl( y \log(p) + (1 - y) \log(1 - p) \bigr),
\end{equation}
where $y_{t,s} = 1$ if $s \in \mathcal{Y}_t^{+}$, and $y_{t,s} = 0$ otherwise. The overall batch objective is then the average over all samples in the batch $\mathcal{S}$.

Because the historical representations $K^{\text{IComp}}_s$, target labels $\mathcal{Y}_t^{+}$, and layer-specific query hidden states $\mathbf{h}_t$ are all pre-extracted and stored offline, the training pipeline achieves complete physical isolation from the host LLM. 
The thousand-billion-parameter backbone model is never loaded into GPU memory during the entire optimization loop. 
Since the trainable projection layers represent less than 0.1\% of the full model's parameter scale, the computational workload is remarkably small. 
As a result, the entire Memory Indexer converges elegantly within \textbf{a single H20 GPU hour}. 

This decoupled design significantly accelerates our research cycle. 
Leveraging a single cluster of 8$\times$ NVIDIA H20 GPUs, we seamlessly executed approximately 500 distinct training runs within a single week to systematically map out the optimal architecture and training strategies, a feat that would be computationally prohibitive under traditional joint end-to-end distillation.

\subsection{Architectural Optimal Configuration}
\label{subsec:optimal_config}

A fundamental premise of designing LSA is that not every transformer layer is suited for contextual lookahead prediction. 
Our early-stage exploration revealed that deploying memory indexers on the initial shallow layers of the LLM yields exceptionally poor lookahead performance, as these early representations predominantly capture low-level token statistics rather than long-range semantic dependencies. 
Therefore, an efficient system routing paradigm must selectively place indexers only on layers that possess mature global context awareness.

However, scaling the number of joint training layers introduces a strict trade-off between performance and serving efficiency. 
While a single-layer retriever lacks the representative capacity to handle diverse long-context workloads, aggressively scaling to an 8-layer joint configuration (spanning layers 6 to 20) introduces severe hardware-side efficiency degradation. 
As verified in our full-system benchmarks, an 8-layer ensemble triggers an excessively loose context recall mask, fetching up to 30\%--49\% of historical compressed KV entries into the GPU memory, which defeats our primary goal of minimizing the memory tax and negating the GPU offload benefit described in Section~\ref{sec:infra}.

Through extensive Pareto-frontier optimization, we established that placing independent Memory Indexers on exactly three strategic intermediate layers---\textbf{layers 10, 12, and 20}---delivers the ultimate sweet spot. 
During inference, our runtime system aggregates the scoring predictions from these three layers using a \textbf{union operations strategy (OR-mode routing)}. 
Specifically, a preceding compressed KV entry is actively fetched into the GPU memory if \textit{at least one} of the three layer indexers predicts its classification score $I_{t,s} \ge 0.5$:
\begin{equation}
    \label{eq:union}
    C_t^{\text{MemComp}} = \bigcup_{l \in \{10, 12, 20\}} \left\{ C_s^{\text{Comp}} \;\middle|\; I_{t,s}^{(l)} \ge 0.5 \right\}.
\end{equation}
This 3-layer consensus framework provides an exceptionally robust fallback protection boundary. 

Our final production model instantiation is built upon this optimal 3-layer geometry and optimized via a carefully curated combination of effective training strategies:
\begin{itemize}
    \item \textbf{Random Initialization:} Rather than loading alignment-biased weights from a host checkpoint, we initialize the indexer's projection matrices randomly, forcing the dual-encoder to learn unified representations from scratch.
    \item \textbf{Query Low-Rank Conditioning:} We leverage the native low-rank query projection geometry of the DeepSeek-V4 architecture. In DeepSeek's MLA/MQA design, the query vector is projected through an internal low-rank bottleneck (officially designated \texttt{q\_lora\_rank} in the DeepSeek-V3 codebase, where the default is 1536). In our implementation, we set this internal projection dimension to $r=2048$ for the R-series configuration. \textbf{This is not PEFT-style LoRA fine-tuning} (which typically uses ranks of 8--64 to learn small perturbations on frozen weights); rather, it is a fixed architectural dimension of the model's attention backbone that determines the representational capacity of the query encoder. Increasing this rank directly expands the spatial projection capacity of the lookahead indexer without introducing any adapter overhead.
    \item \textbf{Focal Loss Denoising:} To prevent easy negative samples from dominating the gradients, we replace standard BCE with a sample-weighted Focal Loss. Let $p_{t,s} \in [0,1]$ denote the Sigmoid-activated indexer score and $y_{t,s} \in \{0,1\}$ the binary label. We first compute the predicted confidence on the correct class:
    \begin{equation}
        p_{t,s}^{\text{(correct)}} = p_{t,s} \cdot y_{t,s} + (1 - p_{t,s}) \cdot (1 - y_{t,s}).
    \end{equation}
    The per-sample Focal Loss is then defined as:
    \begin{equation}
        \mathcal{L}_{\text{FL}} = \frac{1}{|\mathcal{S}|} \sum_{s \in \mathcal{S}} w_{t,s} \, \bigl(1 - p_{t,s}^{\text{(correct)}}\bigr)^{\gamma} \, \ell_{\text{BCE}}(I_{t,s}, y_{t,s}),
    \end{equation}
    where $\mathcal{L}_{\text{BCE}}$ is the standard binary cross-entropy, $\gamma = 2$ is the focusing parameter that down-weights well-classified samples, and $w_{t,s}$ is a per-sample weight. Notably, \textbf{we do not use a separate class-balancing coefficient $\alpha$}; instead, class imbalance is handled jointly by (i) a negative sampling ratio of 3:1 (three negatives per positive) and (ii) the per-sample weight $w_{t,s}$ computed by the \texttt{--weighted-loss} scheduler. This design forces the optimizer to concentrate on hard boundary tokens while keeping the hyperparameter surface minimal.
\end{itemize}

Conversely, multiple popular retrieval and contrastive learning tricks proved to be redundant or even detrimental during our 500-run sweep, and were systematically excluded from our final pipeline:
\begin{itemize}
    \item \textbf{Pairwise-to-Pointwise Chaining:} Transitioning optimization from a pairwise ranking stage (BPR/Margin Loss) to a pointwise calibration stage yielded no statistical recall gains over a pure pointwise training loop.
    \item \textbf{Strong Negative Mining:} Utilizing LLM-annotated semantic chunks as a hard negative pool introduced severe secondary label noise into the contrastive format; random negative sampling within the non-voted historical repository proved significantly more robust.
    \item \textbf{Weighted Loss Functions:} Scaling the loss according to native layer matching counts increased raw precision slightly but degraded the absolute recall bound by discarding boundary context, shifting the model away from its safety-net objective.
\end{itemize}

\paragraph{Note on Hyperparameter Selection.} Due to the unexpected suspension of this project, we were unable to conduct systematic ablation studies on several key hyperparameters. Specifically, the decoding interval $\tau = 64$ and the classification threshold of $0.5$ were selected based on initial exploratory runs but remain untested across alternative values. The 3-layer configuration (layers 10, 12, 20) was determined through the 500-run Pareto sweep described in Section~\ref{subsec:optimal_config}; however, a more fine-grained layer-wise ablation would be desirable for future work.
% =====================================================================
% Infra.tex — subsection of Methodology (Sec 2.5)
% PD-disaggregated serving.
% =====================================================================

\subsection{PD-Disaggregated Serving}
\label{sec:infra}

\begin{figure}[t]
  \centering
  \includegraphics[width=\linewidth]{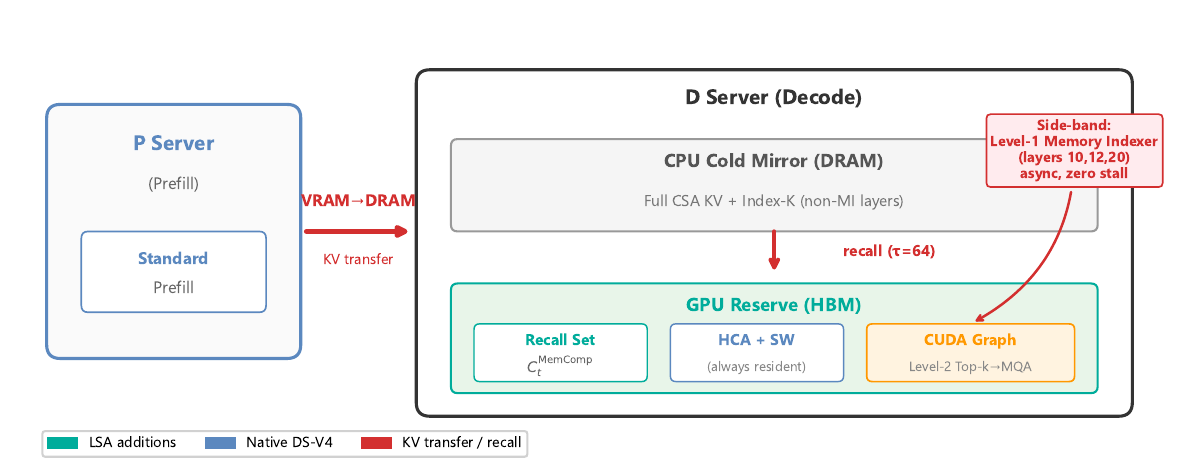}
  \caption{\textbf{PD-disaggregated serving architecture.} P-server streams KV into D's CPU cold mirror (VRAM$\to$DRAM). Level-1 recall runs out of the CUDA graph in an async side band; the graph contains only Level-2 Top-$k$ and MQA over the resident set.}
  \label{fig:infra}
\end{figure}

Realizing LSA's memory saving requires physically evicting non-recalled KV entries from the GPU. We deploy LSA in a prefill--decode (PD) disaggregated serving stack (Figure~\ref{fig:infra}), with all engineering effort on the D-server; the P-server is unmodified DS-V4-Flash.

\paragraph{Prefill and KV transfer.} The P-server runs standard prefill and exports the full KV cache directly into D's host pinned memory (VRAM$\to$DRAM), bypassing D's GPU. Only the always-resident tiers (detailed below) land on D's GPU at admission.

\begin{figure}[t]
  \centering
  \includegraphics[width=\linewidth]{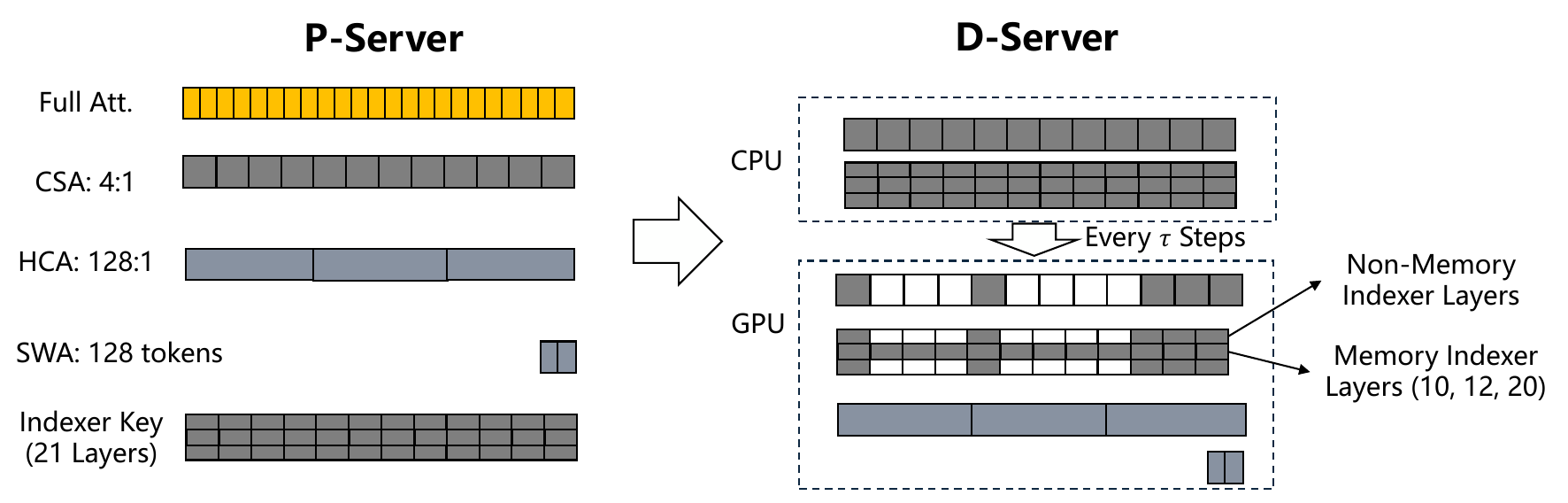}
  \caption{\textbf{KV layout on P-server and D-server.} P-server computes all KV types in full. On D-server, the GPU resident set comprises: (1)~the recalled CSA subset $C^{\mathrm{MemComp}}_t$ (white blocks), (2)~HCA layers, (3)~the last 8K prompt tokens' CSA (sliding window), (4)~all decoded tokens' CSA, and (5)~index-K for MI layers (10, 12, 20). All other CSA KV and non-MI index-K reside in the CPU cold mirror.}
  \label{fig:kv_layout}
\end{figure}

\paragraph{Decode-side offload.} Two categories of KV are offloaded from the GPU to a CPU cold mirror (Figure~\ref{fig:kv_layout}): (1)~non-recalled CSA KV, with only the recalled subset $C^{\mathrm{MemComp}}_t$ (bounded by $K_{\max}$) retained on GPU; and (2)~indexer keys for the 18 non-target layers, with only the 3 Memory-Indexer layers' index-K kept on GPU for scoring. The GPU resident set thus comprises four always-resident tiers---HCA layers (128:1 compression), the CSA chunks corresponding to the last 8K tokens of the original prompt (sliding window), all actively decoded tokens' CSA chunks, and the index-K for layers $\{10,12,20\}$---plus the dynamically recalled CSA subset $C^{\mathrm{MemComp}}_t$. Everything else lives in the CPU cold mirror. The Memory Indexer first fires at decode step 0 (before which all chunks default to resident---byte-identical to baseline); thereafter, every $\tau$ steps the three Memory Indexers score the full history via OR-mode union (Eq.~\eqref{eq:union}) and recall selected pages into the GPU reserve, with zero CPU$\leftrightarrow$GPU transfers within each window.

\paragraph{CUDA-graph-compatible recall.} Level-1 recall involves host-synchronizing operations that prevent CUDA-graph capture. We split LSA across the capture boundary: the captured graph contains only fixed-shape GPU work (read residency bitmap, mask, Top-$k$, remap, MQA), while all Level-1 scoring and copying runs out-of-graph in a side band---approximately $50\times$ faster than the non-CUDA-graph version.

\paragraph{Async recall.} At large batch sizes, stalling inference to wait for recalled KV would severely degrade throughput. We therefore materialize recalled entries at step $t{+}1$ rather than $t$, exploiting the one-step lookahead from the future-window prediction (Eq.~\eqref{eq:label}) to achieve zero decode stall.

Because the offloaded history occupies no GPU KV, the per-request footprint drops from $O(\text{context})$ to $O(|C^{\mathrm{MemComp}}_t|)$. 

\section{Experiments}
\label{sec:experiments}

\subsection{Experimental Setup}
\label{subsec:setup}

To ensure a rigorous and controlled evaluation of the FlashMemory paradigm, we benchmark our model against three structural variants. Crucially, to maintain architectural consistency, \textbf{all evaluated configurations universally retain the full Heavily Compressed Attention (HCA) layers (at a 128:1 compression ratio), alongside the exact CSA chunks corresponding to both the last 8K tokens of the original prompt and all actively decoded tokens within the local window.} The precise treatment of the remaining historical long-context CSA chunks differentiates the methods as follows:

\begin{itemize}
    \item \textbf{\texttt{DS-V4-Flash}}: The standard, unaltered DeepSeek-V4-Flash model.
    \item \textbf{\texttt{FM-DS-V4} (Ours)}: The \texttt{DS-V4-Flash} backbone augmented with the Memory Indexer. The lookahead selection mechanism triggers periodically every $\tau = 64$ decoding steps, dynamically evaluating and fetching query-critical historical CSA chunks from the CPU cold pool into the active GPU HBM.
    
    \item \textbf{\textit{Recency Only}}: A sliding-window fallback control. While it shares the same base HCA layers and the local 8K/decoded CSA window to match the static local memory allocation budget, it completely discards all prior long-context historical CSA chunks and executes zero predictive lookahead retrieval.
    
    \item \textbf{\textit{Random 10\%}}: A naive sparse routing control. On top of the foundational HCA layers and the local 8K/decoded CSA window, it randomly selects and retains exactly 10\% of the global historical context CSA chunks in the active KV cache, providing a non-predictive stochastic baseline.
\end{itemize}

\subsection{Primary Results: Breaking the Capacity Wall}
\label{subsec:main_results}

Table~\ref{tab:primary_benchmarks_transposed_gb} highlights the performance and hardware footprint scaling across three major long-context benchmarks: LongBench-v2~\cite{bai2025longbenchv2deeperunderstanding}, LongMemEval~\cite{wu2025longmemevalbenchmarkingchatassistants}, and RULER~\cite{hsieh2024rulerwhatsrealcontext}.

\begin{table}[htbp]
\centering
\caption{System performance and physical KV cache footprints (GPU memory overhead in gigabytes [GB] in parentheses) across primary long-context benchmarks. \texttt{DS-V4-Flash} operates at 100\% full KV cache allocation without chunk pruning.}
\label{tab:main}\label{tab:primary_benchmarks_transposed_gb}
\begin{tabular}{lcccc}
\hline
\textbf{Benchmark / Dataset} & \textbf{DS-V4-Flash} & \textbf{FM-DS-V4} & \textit{Recency Only} & \textit{Random 10\%} \\ \hline
\textbf{LongBench-v2-S} (46K)   & 68.9 (0.17 GB) & \textbf{70.2} (0.04 GB) & 50.0 (0.03 GB) & 53.3 (0.04 GB) \\
\textbf{LongBench-v2-M} (179K)  & 67.6 (0.65 GB) & \textbf{68.9} (0.08 GB) & 54.4 (0.03 GB) & 48.9 (0.09 GB) \\
\textbf{LongBench-v2-L} (493K)  & 68.1 (1.80 GB) & \textbf{70.0} (0.18 GB) & 54.3 (0.04 GB) & 46.9 (0.22 GB) \\ \hline
\textbf{LongMemEval-S} (125K)   & 80.6 (0.46 GB) & \textbf{82.0} (0.06 GB) & 19.2 (0.04 GB) & 20.1 (0.07 GB) \\
\textbf{LongMemEval-M} (500K)   & 39.3 (1.82 GB) & \textbf{40.2} (0.17 GB) & 23.1 (0.04 GB) & 25.7 (0.22 GB) \\ \hline
\textbf{RULER} (64K)            & 94.7 (0.23 GB) & \textbf{95.0} (0.04 GB) & 36.6 (0.03 GB) & 52.8 (0.05 GB) \\
\textbf{RULER} (128K)           & \textbf{94.3} (0.47 GB) & 93.2 (0.06 GB) & 21.6 (0.03 GB) & 32.3 (0.08 GB) \\
\textbf{RULER} (256K)           & \textbf{90.5} (0.94 GB) & 88.2 (0.09 GB) & 20.6 (0.04 GB) & 41.2 (0.12 GB) \\
\textbf{RULER} (512K)           & 88.3 (1.87 GB) & \textbf{89.6} (0.18 GB) & 18.8 (0.04 GB) & 27.2 (0.22 GB) \\ \hline
\textbf{Avg.}                   & 76.9 (0.93 GB) & \textbf{77.5} (\textbf{0.10 GB}) & 33.3 (0.04 GB) & 38.7 (0.12 GB) \\ \hline
\end{tabular}
\end{table}

The empirical findings deliver a striking victory for the FlashMemory paradigm. Averaged across all tasks, \textbf{\texttt{FM-DS-V4} consumes merely 13.5\% of the baseline GPU memory footprint—representing an average 86.5\% reduction in KV cache storage—while actually improving overall performance to 77.5\% (+0.6\% absolute margin over DS-V4-Flash).} When the average context length reaches 500K, this reduction ratio further climbs to an astonishing 90\%.

This counter-intuitive ``less is more'' phenomenon is especially pronounced in the ultra-long \textbf{LongBench-v2-L (493K)} setting, where our model beats \texttt{DS-V4-Flash} by \textbf{+1.9\%} while running on a threadbare \textbf{10.0\%} memory budget. This forcefully proves our core hypothesis: LSA acts as an expert \textbf{attention denoiser}, filtering out thousands of irrelevant historical chunks that would otherwise clutter the attention dot-products and cause factual hallucinations. Under the same memory restrictions, native heuristic controls (\textit{Recency Only} and \textit{Random 10\%}) completely collapse, failing to synthesize global context and confirming that our indexer has mastered complex predictive temporal routing.

One might naturally question why \textit{Recency Only} and \textit{Random 10\%} can still maintain a reasonable performance baseline on specific datasets like LongBench-v2. It is critical to reiterate that in DeepSeek-V4's hybrid design, the sparse CSA mechanism operates in parallel with the full Heavily Compressed Attention (HCA) layers (at a 128:1 compression ratio). For evaluation scenarios that primarily necessitate global semantic themes or coarse-grained synthesis rather than lossy, hyper-granular token retrieval, utilizing the global compressed HCA foundations alongside the local 8K cache proves sufficient to navigate basic context structures.

\subsection{Limitations and Diagnostics}
\label{sec:limitations}

While FlashMemory achieves unprecedented efficiency gains on three standard long-context benchmarks, our stress-testing exposes critical boundaries of the current paradigm. Due to recent organizational realignments, active development has been suspended. We present these diagnostic findings and concrete failure cases to provide transparent insights for the open-source community.

\subsubsection{Context-Independent Overhead}
\label{subsec:lim_context_independent}
We originally hypothesized that for context-independent queries where historical long context is entirely irrelevant, the pointwise Sigmoid gating would naturally collapse to near-zero retrievals, yielding a strict $O(1)$ constant KV cache footprint. To test this adversarial boundary, we augmented LongMemEval-S and LongMemEval-M by explicitly appending queries that are strictly context-free or tightly bounded to the local 8K window only.

\begin{table}[htbp]
\centering
\caption{System evaluation under adversarial context-independent tasks (No-Context).}
\label{tab:nocontext_benchmarks_transposed_gb}
\begin{tabular}{lcc}
\hline
\textbf{Context Independent Datasets} & \textbf{DS-V4-Flash} & \textbf{FM-DS-V4} (\textbf{Ours}) \\ \hline
\textbf{LongMemEval-S} (No-Context) & \textbf{96.7} (0.46 GB) & 95.0 (\textbf{0.06 GB}) \\
\textbf{LongMemEval-M} (No-Context) & 91.2 (1.82 GB) & \textbf{92.5} (\textbf{0.16 GB}) \\ \hline
\end{tabular}
\end{table}

As shown in Table~\ref{tab:nocontext_benchmarks_transposed_gb}, while the downstream accuracy gracefully matches the foundation baseline, the model \textbf{fails to preserve a constant memory overhead}. Moving from the 125K context to the 500K context, the lookahead memory allocation ratio does scale down to 8.4\%, yet the physical absolute chunk retention volume inflates by approximately \textbf{2.5$\times$}. This indicates that the point-wise Sigmoid gater still leaks a marginal background probability across massive sequence lengths, accumulating false-positive retrievals when facing massive distraction pools.

\subsubsection{Dense Global Memory Breakdown (The MRCR Failure Case)}
\label{subsec:lim_mrcr}
Our model experiences a severe breakdown on the Multi-Range Context Retrieval (MRCR)~\cite{vodrahalli2024michelangelolongcontextevaluations} benchmark, where accuracy plummets from the baseline's 76.0\% down to a \textbf{dismal 48.0\%}. To isolate the root cause of this severe performance regression, we conducted a rigorous oracle simulation: we pre-computed the global golden attention weights of \texttt{DS-V4-Flash} across the full decoding path for each sample, sorted the historical blocks based on cumulative attention density, and selectively loaded only the Top 50\%, 25\%, or 10\% highest-weighted chunks into core MQA layers.

Our diagnostic oracle sweeps revealed a fundamental property difference between benchmarks: for LongBench-v2, LongMemEval, and RULER, retaining a mere 10\% or 25\% of golden CSA chunks alongside global HCA layers completely secures 100\% baseline accuracy. However, MRCR exhibits an aggressive \textbf{global dense memory dependency}—even when providing the indexer with 50\% of the absolute \textbf{true golden chunks}, the accuracy still drops by about 2\% compared to full-context cache execution. 

These two empirical findings firmly isolate the architectural limitations of our current Memory Indexer. Ideally, we envisioned an ideal indexer capable of executing deterministic, context-adaptive retrieval: achieving near-zero recall on context-independent tasks to maintain a constant memory floor, while delivering near-perfect recall on memory-dense tasks to secure maximum contextual awareness. 

Unfortunately, by relying on a highly compressed, standalone Dual-Encoder framework, the model fundamentally lacks the capacity to balance such extreme operational boundaries of precision and recall. Consequently, the following three critical factors bound its performance:
\begin{enumerate}
    \item \textbf{Frozen Key Representation}: Due to computational budget constraints, we never adjusted or optimized the native DeepSeek-V4 Compressed indexer keys ($K^{\text{IComp}}$), fine-tuning only the query projection encoder.
    \item \textbf{Shallow Cross-Interaction}: Operating purely via a 64-step coarse dot-product similarity, the indexer lacks the multi-turn interaction capacity. Incorporating a \textbf{Late-Interaction architecture} (e.g., ColBERT-style token-level cross-matching) is essential to untangle complex dense retrieval patterns.
    \item \textbf{Decoupled Training Isolation}: The lack of end-to-end joint optimization with the main backbone restricts the indexer to static pseudo-labels, ignoring live autoregressive shift dynamics. 
\end{enumerate}
Addressing these items remains our formal future roadmap.

\subsubsection{The Length Generalization Ceiling}
\label{subsec:lim_length}
Our initial design intent assumed that because our lookahead indexer operates via point-wise chunk matching, we could train the Dual-Encoder on relatively short context windows (e.g., 128K) and seamlessly scale zero-shot inference to 1M+ context fields, as candidate pool expansion theoretically shouldn't distort point-wise scoring.

Our empirical evaluations completely dismantled this assumption. The indexer safely generalizes \textbf{up to exactly $2\times$ its training context length}. Attempting to execute inference beyond this hard boundary causes accuracy to collapse precipitously, with lookahead block selection degenerating into near-random sampling. We attribute this performance bottleneck to the effects from the out-of-distribution positional embeddings, which constitutes the primary architectural divergence between self-attention mechanisms and generic text retrieval systems. Consequently, our final released memory indexer was explicitly trained on context lengths up to 512K. Although empirical validation at greater scales remains untested, we hypothesize that its retrieval discriminability would decay irreversibly when deployed on sequences exceeding 1M tokens.

\section{Conclusion}
\label{sec:Conclusion}

In this report, we have presented FlashMemory-DeepSeek-V4, an LLM augmented with Lookahead Sparse Attention (LSA). By introducing a Neural Memory Indexer into the DeepSeek-V4-Flash architecture, we enable the model to proactively predict and fetch only the query-critical KV chunks into GPU memory. Compared to DeepSeek-V4-Flash, our model achieves comparable or even superior performance across the majority of benchmarks, while consuming merely approximately 13.5\% of the GPU memory. At 1M context, this translates to a 0.30$\times$ per-token compute ratio, 90\% KV cache reduction, and \textbf{2.8$\times$ aggregate throughput with 2.7$\times$ concurrency gains} in PD-disaggregated serving.

We emphasize that the architecture, training pipeline, and hyperparameters of FlashMemory-DeepSeek-V4 are severely constrained by computational resources and the unexpected suspension of the project. The indexer was trained with frozen key representations, shallow dot-product interaction, and no end-to-end joint optimization with the backbone—design choices dictated by resource availability rather than optimality. Nevertheless, the results achieved under these constraints make us highly confident in the vast potential for improvement that remains: FlashMemory-DeepSeek-V4, in its current form, is merely the first glimpse of what LSA can achieve for ultra-long-context intelligence.

\bibliographystyle{unsrt}
\bibliography{custom}

\end{document}